# Earthquaker-AI: A Retrieval-Augmented Generation Framework with Rubric-Based Assessment for Primary School Earthquake Education

Xanthi Kokkinou[1], Chaido Mizeli[2], Nafsika Koulaxidou[3], Marina Delianidi[4], Konstantinos Diamantaras[5]

[1]Primary Education of Larissa, Larissa, Greece — kokkinouxa@gmail.com
[2]International Hellenic University, Thessaloniki, Greece — cmizeli@ihu.edu.com
[3]Primary Education of Larissa, Larissa, Greece — nausikak1966@gmail.com
[4]International Hellenic University, Thessaloniki, Greece — dmarina@ihu.gr
[5]International Hellenic University, Thessaloniki, Greece — kdiamant@ihu.gr

**Abstract—**This paper presents Earthquaker-AI, a hybrid educational framework that builds upon a previously implemented educational robotics project by integrating a conversational artificial intelligence assistant based on Retrieval-Augmented Generation (RAG), aiming to enhance earthquake preparedness and conscious action among primary-school students. The system extends the earlier award-winning STEM project Earthquaker, which received 3rd place in the Hellenic WRO Competition (2020), and moves from mechanical simulation through Lego WeDo2 activities to cognitive and metacognitive processing.

The robotics component, developed in the original Earthquaker project, employed Lego WeDo2-based automation to simulate seismic response, allowing students to interact with sensors and actuators that function as tangible representations of core protective actions. At the cognitive level, the conversational assistant adopts a Retrieval-Augmented Generation architecture, in which evidence retrieval precedes language generation, so that responses are grounded in official educational material and aligned with the intended pedagogical framework. The assistant serves a dual role: it operates as a guided learning mechanism that aligns students' responses with institutional safety guidelines, while simultaneously providing rubric-based verbal feedback that supports self-regulated learning and the development of calmness under emergency conditions.

Earthquaker-AI follows a progressive learning trajectory aligned with students' cognitive development across age stages. In the early grades, the emphasis is placed on basic recognition of earthquake-safety actions through simple multiple-choice questions, in which students select the most appropriate answer, supporting orientation toward the recommended behavior during seismic events. Assessment at this stage is conducted using a two-dimensional rubric focused on action recognition and emotional regulation. In middle grades, learning activities become more demanding, requiring students to identify the correct sequence of actions through multiple-choice questions, evaluated using a three-axis rubric that captures organized thinking and decision-making. In upper grades, the approach shifts from recognition to verbal production, with students providing short written responses assessed through a four-dimensional rubric. The additional dimension of clarity of expression reflects increasing metacognitive maturity, as it requires structured reasoning, justification, and precise articulation.

The system also includes a dedicated dialogic module that leverages RAG to retrieve and synthesize evidence-based answers. Student queries are semantically matched with selected excerpts from official earthquake-safety guidelines, from which pedagogically safe and accurate responses are generated. Experimental evaluation demonstrates high answer groundedness (0.84) and accuracy (0.85), together with a low hallucination rate (0.07), indicating stable and evidence-aligned behavior under the evaluated experimental conditions.

Overall, Earthquaker-AI presents an integrated educational approach to earthquake preparedness, bringing together hands-on engagement, information processing, and reflective practice. The sequence of observation, physical interaction, interpretation, and verbal articulation provides a holistic learning experience. At the same time, the combined use of robotic processes, analytic rubrics, and artificial intelligence promotes technological literacy, self-regulation, and responsible use of digital systems in primary education, contributing meaningfully to earthquake preparedness and the early development of crisis-management skills.

## 1. Introduction

Earthquake preparedness constitutes a critical component of safety education in primary schools. Despite the existence of official guidelines, earthquake education at the primary level often remains fragmented and predominantly theoretical, offering limited opportunities for experiential learning, structured reflection, and individualized feedback (Yeon et al., 2020). While STEM-oriented activities and educational robotics have been successfully employed to promote engagement and hands-on understanding of physical safety mechanisms, their impact frequently remains confined to mechanical simulation and does not systematically extend to the cognitive and verbal articulation of appropriate actions during an earthquake. Comparative research on earthquake training methods further indicates that interactive and game-based approaches can enhance engagement and learning outcomes compared to traditional instruction (Çoban & Göktaş, 2022).

The original Earthquaker project addressed the limitations of predominantly theoretical earthquake education by introducing a robotics-based STEM framework focused on the design and implementation of an automated school safety system for earthquakes. Inspired by ancient automation mechanisms, the original Earthquaker system enabled students to simulate seismic detection and automated responses such as door opening, utility shutdown, and occupancy counting. Through hands-on construction and physical interaction, students developed an experiential understanding of cause–effect relationships related to earthquake safety. However, while the robotic component effectively supported procedural awareness, it offered limited support for guided reasoning, justification of actions, and formative assessment of students' understanding.

Recent work has highlighted both the opportunities and challenges of integrating large language models into educational settings (Kasneci et al., 2023). In this context, Earthquaker-AI emerges as a natural evolution of this framework, extending earthquake education from the domain of mechanical causality to the cognitive and metacognitive dimensions of preparedness. The transition to Earthquaker-AI responds to the identified need for structured, evidence-based, and pedagogically controlled learning experiences that support not only correct actions, but also calmness, judgment, and verbal expression during emergency situations. To address this need, the system integrates experiential robotics with a dialogic artificial intelligence component designed to support scenario-based learning and structured evaluation.

Earthquaker-AI integrates a Retrieval-Augmented Generation (RAG) architecture within an educational setting. By design, all system responses are generated exclusively on the basis of retrieved excerpts from official earthquake safety guidelines, ensuring that learning interactions remain accurate, evidence-grounded, and constrained within verified instructional sources. This design aims to reduce risks typically associated with open-domain conversational agents and allows the system to function as a controlled educational assistant rather than a generic chatbot.

In parallel, Earthquaker-AI incorporates rubric-based formative assessment as a core pedagogical mechanism. Student interactions are structured according to age-appropriate scenarios, and responses are evaluated using explicitly defined rubrics aligned with students' cognitive development. This design supports pedagogical continuity across educational levels and enables the provision of targeted, comprehensible feedback that fosters self-regulation and reflective learning.

The objectives of this study are threefold: (a) to present the design of Earthquaker-AI as a hybrid educational model situated within an existing robotics-based learning environment and extended with RAG-based dialogue, and rubric-driven assessment; (b) to evaluate the extent to which the RAG component produces accurate and evidence-aligned responses grounded in official earthquake safety guidelines; and (c) to examine the reliability and consistency of rubric-level formative assessment produced by language models across repeated evaluations and different educational grades.

## 2. Conceptual Positioning of Earthquaker-AI

Earthquaker-AI represents a natural pedagogical extension of the original STEM project Earthquaker: Design and Implementation of an Automated School Safety System for Earthquakes, transferring the logic of mechanical causality and physical simulation into a broader cognitive and pedagogical framework. In the initial project, students engaged in robotic constructions that simulated seismic detection and automated response mechanisms (e.g., door opening, utility shutdown, and occupancy counting), enabling them to develop an experiential and procedural understanding of fundamental safety principles. The transition toward Earthquaker-AI does not replace this experiential level; rather, it builds upon it, shifting the emphasis from physical execution to cognitive processing, verbal articulation, and informed decision-making, as highlighted in the system's design.

The need for this transition is grounded in an identified pedagogical gap: earthquake education in primary schools often remains largely theoretical and fragmented, with limited opportunities for individualized feedback. Despite the existence of official guidelines, students are rarely invited to actively process earthquake-related scenarios, justify their choices, or be assessed using clear, age-appropriate, and comprehensible criteria. Earthquaker-AI seeks to address this gap by relocating earthquake preparedness into the domain of guided inquiry and gamified learning, where knowledge is not passively reproduced but actively constructed through structured interaction.

## 3. Pedagogical and Developmental Foundations

The concept of pedagogical continuity (learning continuum) is established as a central theoretical axis of Earthquaker-AI. The system is designed as a progressive learning trajectory, aligned with students' cognitive and metacognitive development, in accordance with principles of developmental psychology (Piaget, 1970) and the spiral organization of the curriculum (Bruner, 1960). Learners are not approached as a homogeneous group; rather, differentiated cognitive processing capacities are assumed, necessitating the systematic gradation of cognitive demands and assessment tools. The robotic component of Earthquaker-AI is grounded in constructionist design principles, emphasizing learning through hands-on manipulation and iterative exploration (Resnick & Rosenbaum, 2013).

Within this framework, assessment rubrics function as the structural mechanism linking learning and assessment. As a formative assessment instrument, performance criteria and quality levels are made explicit, allowing expectations and evaluation procedures to be transparent and comprehensible to students (Andrade, 2000; Brookhart, 2013). In Earthquaker-AI, the use of rubrics is framed as non-punitive and non-comparative, with emphasis placed on supportive application and the provision of targeted, pedagogically adapted feedback.

The differentiation of rubrics across grade levels is aligned with students' cognitive development. In the early grades (Grades 1–2), assessment focuses on the recognition of basic actions and the maintenance of calm behavior, with reduced cognitive load and explicit support for self-regulation in crisis situations (Zimmerman, 2002). In the middle grades (Grades 3–4), the inclusion of action sequencing enables the assessment of organized thinking and systemic understanding, both of which are considered essential for earthquake preparedness and for integration with the project's robotics component. In the upper grades (Grades 5–6), emphasis is placed on verbal production, justification, and clarity of expression, which are treated as indicators of metacognitive maturity and conscious action (Flavell, 1979; Panadero & Jonsson, 2013). The feedback mechanism of Earthquaker-AI aligns with established models of formative feedback that support self-regulated learning and performance improvement (Hattie & Timperley, 2007).

This design choice is aligned with contemporary literature that highlights rubrics as powerful tools for self-assessment and learning regulation (Andrade & Du, 2005; Brookhart, 2017). At the same time, it enables the artificial intelligence system to operate not as an authoritative evaluator but as a pedagogical mediator, in which students' responses are compared against institutional standards, and feedback is returned in a comprehensible and actionable form.

Overall, the theoretical framework of Earthquaker-AI is articulated as a hybrid pedagogical model that integrates experiential robotics, structured knowledge, and formative assessment into a coherent learning scheme. The transition from mechanical simulation to cognitive processing, the use of Retrieval-Augmented Generation (RAG) for pedagogically controlled dialogue, and the staged application of rubrics collectively constitute a framework oriented not only toward the acquisition of correct actions, but toward conscious understanding, calmness, and students' metacognitive awareness in earthquake-risk situations.

## 4. AI & RAG Framework

A central component of the framework is the adoption of a Retrieval-Augmented Generation (RAG) architecture (Lewis et al., 2020). This approach combines natural language generation with semantic information retrieval from external, controlled, and institutionally validated sources. The retrieval component relies on semantic embeddings generated by a Greek-language Sentence Transformer model (dimitriz/st-Greek-media-Bert-base-uncased) and indexed with FAISS, following established retrieval approaches for open-domain question answering (Karpukhin et al., 2020). At the pedagogical level, the use of RAG ensures that generated responses are not derived from generalized or arbitrary model knowledge, but are grounded exclusively in retrieved excerpts from official earthquake safety guidelines. The system is designed to operate within pedagogically controlled constraints, reducing the risk of misinformation and strengthening trust and reliability in the learning process (UNESCO, 2014). In this context, the adoption of RAG can be interpreted as a mechanism of epistemic control, ensuring that generative processes remain strictly bounded by verifiable institutional knowledge rather than unrestricted model priors.

## 5. System Design

Earthquaker-AI is implemented as an educational AI layer designed to support exploratory learning and structured formative assessment within the existing earthquake-preparedness environment. The system combines Retrieval-Augmented Generation (RAG) with rubric-based evaluation and guided pedagogical feedback, ensuring that all interactions remain strictly grounded in official earthquake safety guidelines and aligned with age-appropriate educational objectives.

### 5.1 Interaction Modes

Earthquaker-AI is structured around two complementary interaction modes, each addressing distinct pedagogical goals.

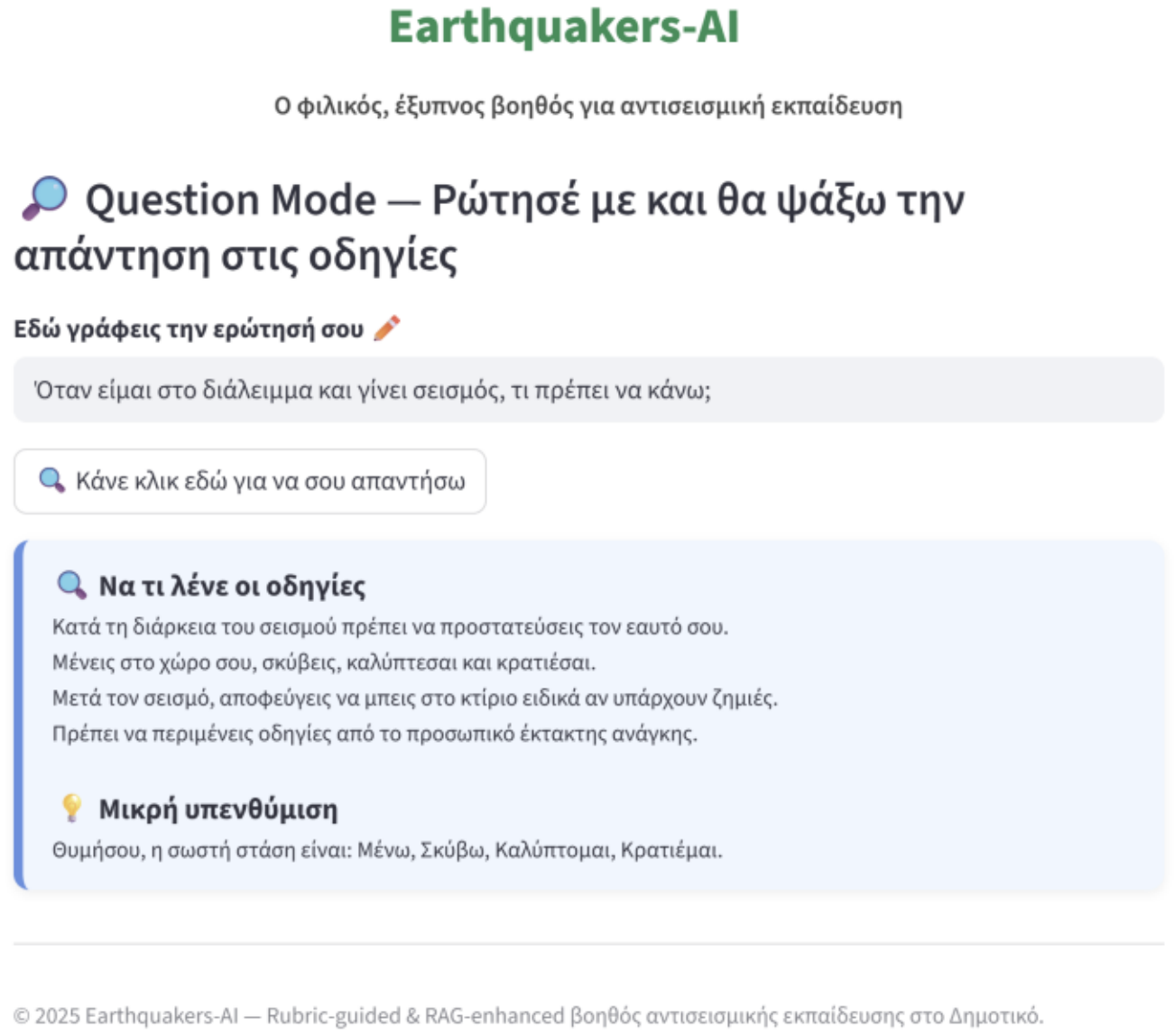

Figure 1. Example of the Question Mode interface, illustrating evidence-grounded answering based on retrieved excerpts from official earthquake-safety guidelines.

Question Mode allows students to freely ask questions related to earthquake safety. For every query, the system retrieves semantically relevant excerpts from the official earthquake safety corpus and generates an answer grounded in the retrieved material. This mode is designed to promote inquiry-based learning, while ensuring safety and factual reliability, since no response can be produced without explicit grounding in authoritative sources. An example of the Question Mode interface is presented in Figure 1.

Quiz Adventure Mode provides scenario-based questions tailored to the student's educational grade. Students respond either through structured selections (e.g., multiple-choice) or through short open-ended answers, which are then evaluated using predefined, grade-specific rubrics. For each response, the system produces rubric-level scores and short, pedagogically adapted feedback, followed by a concise progress summary at the end of each session. This mode integrates learning and assessment, enabling students to reflect on their performance while receiving structured instructional guidance. An example of the Quiz Adventure interaction is shown in Figure 2.

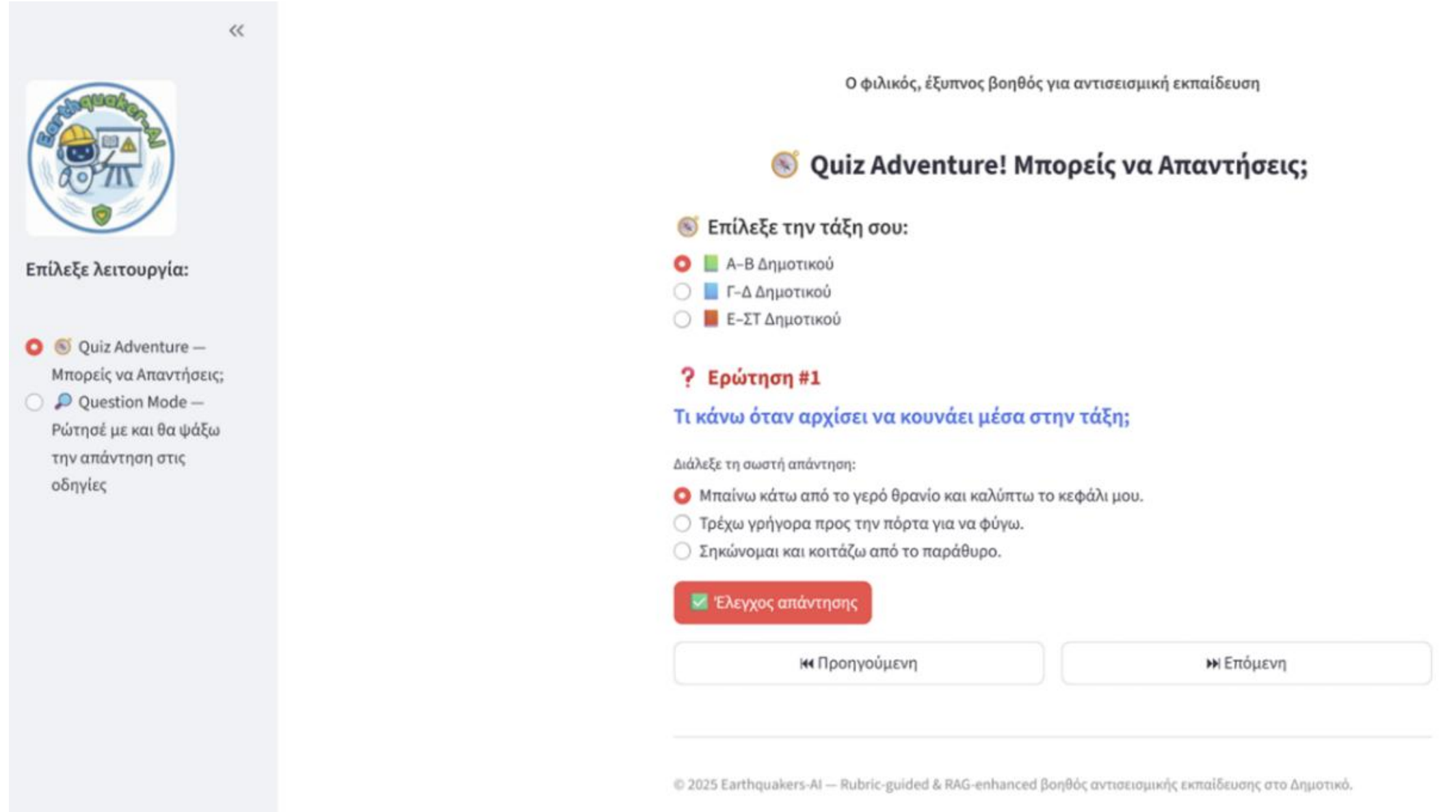

Figure 2. Example of the Quiz Adventure mode (Greek UI), illustrating scenario-based, grade-adapted interaction.

### 5.2 RAG–Rubric Evaluation Pipeline

All interactions in Earthquaker-AI are organized through a multi-stage evaluation pipeline, as illustrated in Figure 3. First, the system loads the question content, its interaction type, and the reference answer from a grade-specific structured repository. Second, it performs semantic retrieval over the official earthquake safety corpus to obtain the most relevant evidence passages, which serve as the basis for both response generation and evaluation. Third, a language model assigns rubric-level scores (1–3) for each assessment axis by comparing the student's response with the reference answer and the retrieved evidence. Fourth, the system generates pedagogically adapted feedback aligned with the assigned rubric levels and highlights strengths and areas for improvement.

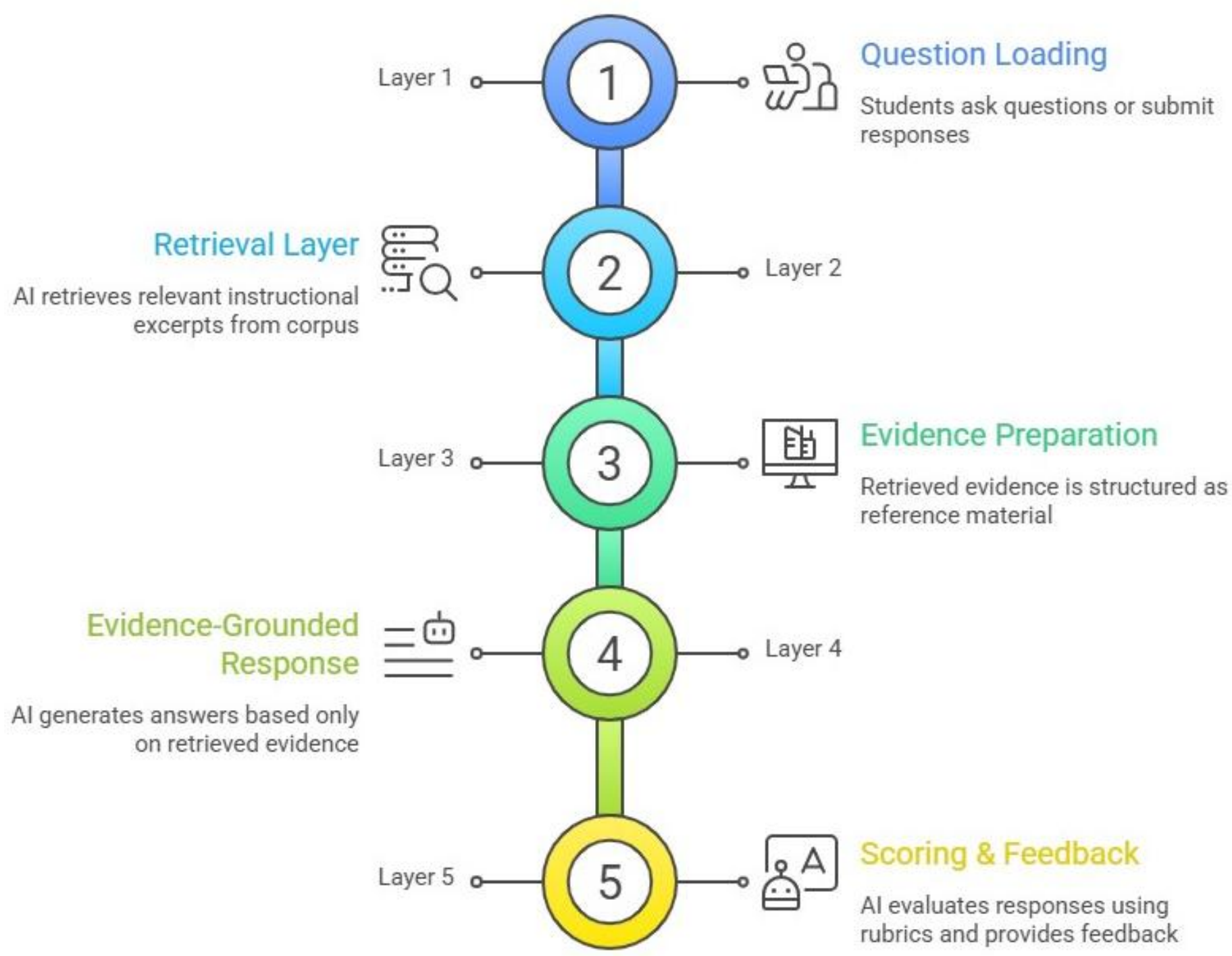

Figure 3. Overview of the Earthquaker-AI system architecture and RAG–Rubric evaluation pipeline

Figure 3 illustrates the overall architecture of the Earthquaker-AI system, including the RAG retrieval process, rubric-based evaluation, guided feedback generation, and final progress summarization. This architecture ensures that Earthquaker-AI does not function as an unrestricted conversational agent, but as a structured educational system, in which all answers, evaluations, and feedback are systematically anchored in verified instructional content and explicit assessment criteria.

Finally, after a set of questions has been completed, a concise, individualized progress summary is produced. An example of rubric-based feedback and the generated progress summary is presented in Figure 4.

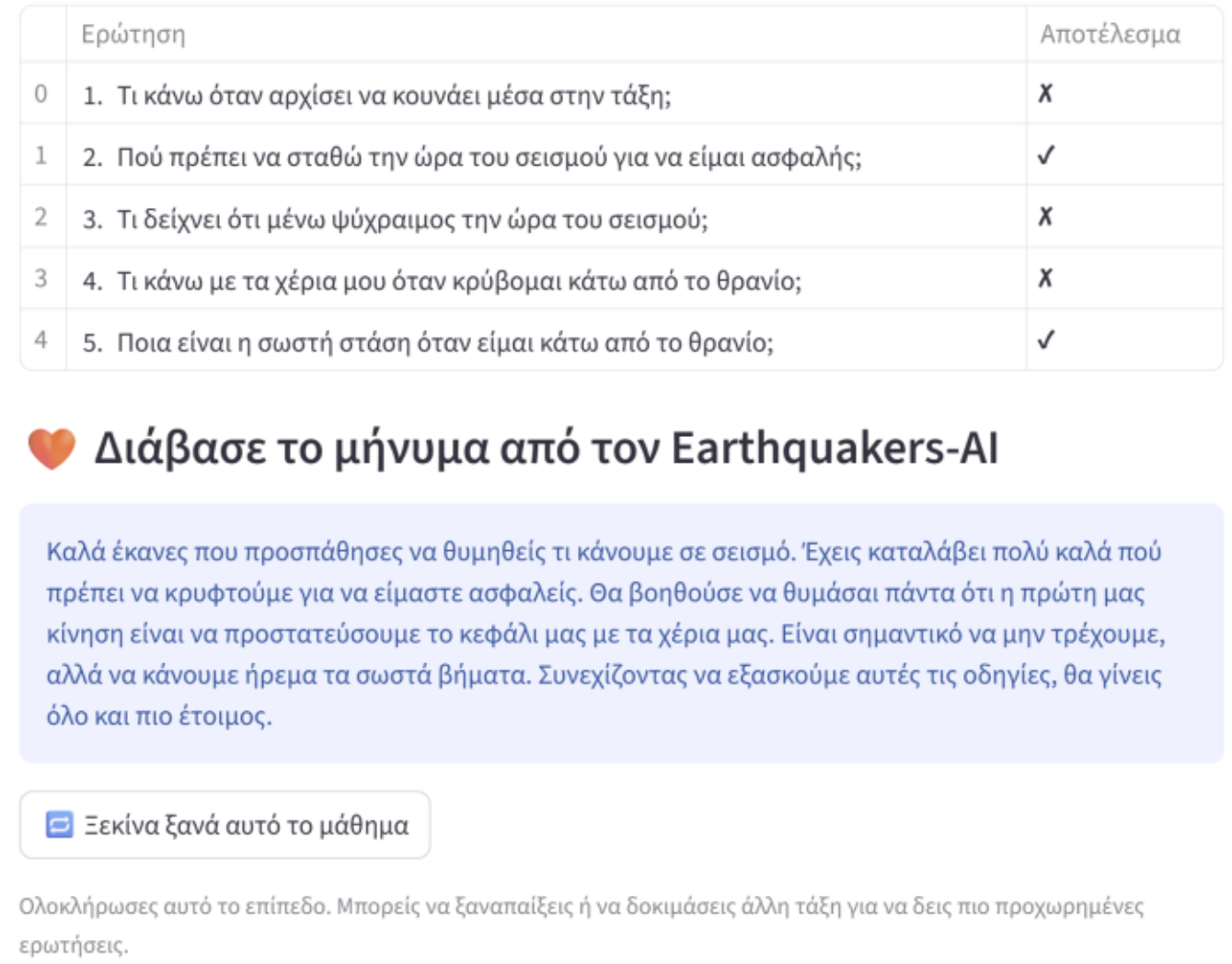

Figure 4. Example of rubric-based evaluation and individualized progress summary displayed to students after completing a question set.

## 6. Research Objectives and Research Questions

The Earthquaker-AI project is designed to support safe and evidence-based earthquake education for primary school students through structured AI-assisted interaction and formative assessment. Accordingly, this study pursues two specific research objectives: (a) to assess the factual accuracy and evidence grounding of answers generated by the RAG component, and (b) to evaluate the consistency and reliability of rubric-based formative assessment produced by AI models.

These objectives are operationalized through the following research questions:
RQ1: To what extent does Earthquaker-AI generate factually accurate and evidence-grounded answers that are explicitly supported by retrieved excerpts from official earthquake safety guidelines?
RQ2: To what extent does the AI-based assessment model (SLM/LLM) produce consistent and accurate rubric-level scores (1–3) under stochastic generation, across repeated evaluations of standardized student responses?

### 6.1 Methodology

The study adopts a two-stage experimental design, directly aligned with the defined research questions. The first stage (RQ1) evaluates the answer quality and grounding behavior of the Retrieval-Augmented Generation (RAG) component, focusing on factual accuracy and alignment with retrieved instructional evidence. The second stage (RQ2) examines the stability and accuracy of rubric-level scoring, comparing repeated evaluations of predefined standardized student responses produced by language models of different sizes.

### 6.2 Evaluation of RAG Answer Quality (RQ1)

To assess the reliability of the RAG component, a set of representative student-style questions was submitted to the system. For each question, the system retrieved semantically relevant excerpts from the official earthquake safety corpus and generated an answer based on these retrieved excerpts. The retrieval corpus remained fixed and unchanged across all experimental runs, ensuring controlled and reproducible evaluation conditions.

The generated answers were evaluated using an LLM-as-a-Judge approach, implemented using an independent Gemma 3 (4B) model. This evaluation strategy follows recent methodological practices in which large language models are employed as judges for assessing the quality and faithfulness of generated outputs (Zheng et al., 2023). The responses being evaluated were generated by the DeepSeek model within the RAG pipeline, while Gemma 3 (4B) was used solely as an independent evaluation model. The judge assessed each answer along four computationally defined criteria:

- Groundedness: the degree to which the answer is explicitly supported by the retrieved excerpts,
- Accuracy: the factual correctness of the answer with respect to the official earthquake safety guidelines,
- Completeness: the extent to which all relevant aspects of the question are adequately covered,
- Hallucination risk: the presence of information not supported by the retrieved evidence.

Each criterion was scored numerically, allowing the computation of aggregate indicators for overall RAG performance. These evaluation dimensions are conceptually aligned with recent automated frameworks for assessing Retrieval-Augmented Generation (Es et al., 2023). In addition, complementary embedding-based similarity metrics were computed between generated answers and retrieved evidence passages in order to provide a system-level, embedding-based assessment independent of the generation model.

### 6.3 Rubric-Based Assessment and Model Comparison (RQ2)

To evaluate the reliability of the rubric-based assessment mechanism, standardized evaluation scenarios were constructed. For each question, three types of predefined student-style responses were constructed, representing high-, medium-, and low-quality answers. Each response was associated with predefined expected rubric levels (1–3) for all assessment axes of the corresponding educational grade.

These scenarios were submitted repeatedly (five runs per scenario) to two language models of different sizes: a Small Language Model (Gemma 3, 4B) and a Large Language Model (DeepSeek). For each run, the model produced rubric-level scores (1–3) based on the student response, the reference answer, and the retrieved evidence.

Performance was evaluated using two primary indicators:

- Assessment accuracy, defined as the percentage of cases in which the predicted rubric level matched the expected level,
- Assessment consistency, measured both as absolute consistency (identical scores across all repetitions) and relative consistency (dominance of the most frequent score across repetitions).

This second stage allows us to examine the stability and internal reliability of AI-based formative assessment, as well as comparative behavior across language models of different capacities, within the Earthquaker-AI framework. In RQ2, rubric-level predictions were directly compared against predefined expected levels; no external LLM judge was used in this stage.

## 7. Experimental Setup

The experimental evaluation was conducted using a set of earthquake-safety questions and structured student-response scenarios, constructed on the basis of the official earthquake safety guidelines. The evaluation included 15 earthquake-safety questions across three grade clusters (A–B, G–D, and E–ST). For RQ1, student-style questions were submitted to the RAG-based system, which retrieved relevant evidence passages and generated grounded answers. This resulted in 15 RAG-generated responses, which were subsequently evaluated for groundedness, accuracy, completeness, and hallucination risk. For RQ2, three standardized student responses (high-, medium-, and low-quality) were created for each question and evaluated repeatedly. Each scenario was executed five times on two language models of different sizes: a Small Language Model (Gemma 3, 4B) and a Large Language Model (DeepSeek). Given 15 questions, three predefined response-quality levels per question, five stochastic repetitions, and two language models, the experimental design comprised 450 rubric-level evaluation instances. This design enabled the measurement of assessment accuracy and evaluation consistency across repeated stochastic runs.

## 8. Results

### 8.1 Results for RQ1 – RAG Answer Quality

To address RQ1 ("To what extent does Earthquaker-AI provide accurate and evidence-aligned answers grounded in retrieved official earthquake safety guidelines?"), the evaluation was conducted at two complementary levels: (a) system-level assessment of retrieval-supported generation within the RAG pipeline, and (b) semantic evaluation of the generated answers.

#### 8.1.1 System-Level Evaluation of RAG Retrieval

As shown in Table 1, system-level computational metrics indicate that the RAG pipeline consistently retrieves semantically relevant excerpts from the official earthquake safety corpus. High similarity between retrieved evidence and generated responses, together with adequate coverage of instructional content, indicates strong alignment with the retrieved source material.

The system-level hallucination_ratio does not measure factual fabrication, but rather the dispersion of answer content across multiple retrieved chunks, serving as an indicator of retrieval distribution rather than unsupported generation. These results indicate that the RAG pipeline provides a reliable retrieval foundation for evidence-grounded answer generation.

| Metric | Mean |
|---|---|
| Support from most relevant retrieved source (sim_best_evidence) | **0.78** |
| Average semantic similarity with retrieved chunks (mean_chunk_similarity) | **0.62** |
| Coverage of key safety points (coverage_ratio) | **0.81** |
| Hallucination indication at system level (hallucination_ratio) | **0.37** |
| Average answer length (tokens) | **65.9** |

Table 1. System-level computational metrics evaluating the behavior of the RAG retrieval pipeline.

### 8.1.2 Semantic Evaluation of Generated Answers

At the semantic level, answer quality was evaluated using an LLM-as-a-Judge approach (Gemma 3, 4B).

| Metric | Mean |
|---|---|
| Groundedness (answer supported by retrieved excerpts) | **0.84** |
| Accuracy (alignment with official guidelines) | **0.85** |
| Completeness (coverage of required information) | **0.78** |
| Hallucination (unsupported information) | **0.07** |

Table 2. LLM-as-a-Judge (Gemma 3, 4B) evaluation of RAG-generated answers across groundedness, accuracy, completeness, and hallucination risk.

As reported in Table 2, the system achieved high groundedness (0.84) and accuracy (0.85), while maintaining a low hallucination rate (0.07). These results indicate that the RAG configuration effectively constrained generation within the retrieved evidence, supporting reliable answer production under the evaluated conditions.

Overall, the results indicate that Earthquaker-AI provides accurate and evidence-grounded answers, addressing the requirements of RQ1.

## 8.2 Results for RQ2 – Rubric-Based Assessment Reliability

The second research question examined the reliability and stability of rubric-level scoring produced by language models of different sizes.

### 8.2.1 Rubric-Level Accuracy

As shown in Table 3, the Large Language Model (DeepSeek) achieved higher rubric-level accuracy than the Small Language Model (Gemma 3, 4B). In particular, the LLM more frequently assigned rubric scores that matched the predefined expected levels for medium- and high-quality response scenarios, while the SLM exhibited lower agreement in these categories. This pattern highlights the influence of model capacity on rubric-level assessment accuracy within the defined experimental setup.

| Model | Rubric-Level Accuracy (%) |
| --- | --- |
| Gemma (SLM) | 54.7 |
| DeepSeek (LLM) | 73.7 |

Table 3. Rubric-level assessment accuracy across models.

### 8.2.2 Assessment Consistency Across Repeated Evaluations

Assessment stability was evaluated using both absolute and relative consistency metrics. As reported in Table 4, both models demonstrated high absolute consistency across repeated evaluations, indicating that identical rubric levels were assigned across repeated runs in the majority of cases. The LLM exhibited slightly higher absolute consistency values.

In addition, Table 5 shows that relative consistency remained very high for both models, with the LLM achieving the highest dominance of the most frequent rubric score across repetitions.

| Model | Absolute Consistency (%) |
| --- | --- |
| Gemma | 87.5 |
| DeepSeek | 89.7 |

Table 4. Absolute assessment consistency across repeated evaluations.

| Model | Relative Consistency (%) |
| --- | --- |
| Gemma | 95.7 |
| DeepSeek | 97.2 |

Table 5. Relative assessment consistency across repeated evaluations.

Overall, these results indicate that while both models provide stable rubric-based evaluations, model capacity appears to influence assessment accuracy, with the LLM consistently outperforming the SLM. At the same time, high consistency values across both models suggest that AI-based rubric evaluation demonstrates reproducible behavior under the defined experimental conditions and when guided by explicit rubrics. These results suggest that while smaller models may provide stable evaluations, higher-capacity models are better suited for fine-grained rubric-level discrimination.

## 9. Ethical, Legal and Data Protection Considerations

Earthquaker-AI was designed in alignment with ethical and legal requirements for educational use in primary school settings. The system does not collect, process, or store personal data of students and operates exclusively on anonymized, scenario-based inputs.

All knowledge sources are restricted to official earthquake safety guidelines retrieved through a closed RAG corpus.

This design ensures that the system does not expose students to uncontrolled or unsafe content, in contrast to open-domain conversational agents. Under current Greek and European regulatory frameworks, the experimental and supervised use of AI systems for educational purposes is permitted, provided that data protection, safety, and pedagogical oversight are ensured.

In addition, the system supports deployment in a fully local execution setting for educational use. In the experimental configuration, the Small Language Model (Gemma 3, 4B) was executed locally via the Ollama framework for the assessment experiments, without any external API calls or data transmission to third-party servers. This local execution design is intended to support GDPR-aligned deployment in primary-school contexts, as all interactions, evaluations, and generated outputs remain confined to the local computational environment. Such architecture is particularly appropriate for primary-school contexts, where minimizing data exposure and ensuring institutional control over educational AI systems is a critical ethical requirement. No real student data were used in the experimental phase; all evaluations relied exclusively on predefined scenario-based inputs.

## 10. Limitations

The present study is subject to certain limitations. The experimental evaluation was conducted in a controlled, scenario-based setting and did not involve real-time interaction with students in classroom environments. In addition, the assessment focused on answer quality and rubric-level consistency, rather than on long-term learning outcomes or behavioral change. These limitations define the scope of the findings and motivate future classroom-scale and longitudinal studies.

## 11. Conclusion

This study presented Earthquaker-AI as a hybrid educational model that builds upon an existing robotics-based learning environment by integrating Retrieval-Augmented Generation (RAG) and rubric-based formative assessment to support earthquake preparedness in primary education. By combining experiential STEM activities with AI-mediated, evidence-grounded dialogue and structured evaluation, the system addresses limitations of purely theory-driven approaches by integrating experiential and dialogic learning components.

The results for RQ1 indicate that, under the defined experimental configuration, the RAG architecture enables the system to generate responses that remain closely aligned with retrieved official earthquake safety guidelines. The observed levels of groundedness and accuracy, combined with low hallucination rates, suggest that constraining generation to verified instructional sources can enhance reliability in safety-critical educational contexts.

These findings highlight the potential of RAG as a pedagogically bounded mechanism for AI-assisted interaction within controlled educational settings.

The results for RQ2 show that AI-based rubric scoring can provide stable and reproducible formative assessment, with model capacity influencing performance. The Large Language Model (LLM) achieved higher rubric-level accuracy than the Small Language Model (SLM), while both models exhibited high consistency across repeated evaluations. These findings indicate that, when guided by explicit rubrics and evidence, language models can support structured educational assessment rather than producing unstructured or inconsistent grading outputs.

Overall, the findings indicate that an evidence-based AI architecture can support structured learning interactions and rubric-based evaluation in primary-school earthquake education. The combination of RAG-grounded dialogue, rubric-driven evaluation, and age-appropriate feedback provides a scalable and transparent framework for the use of artificial intelligence in disaster-preparedness education. In line with recent discussions on both the opportunities and risks of LLMs in education, the present study demonstrates that constraining generative models through evidence retrieval and explicit rubrics can support pedagogically controlled deployment. Future work will focus on classroom-scale experimentation and the longitudinal examination of learning outcomes across different educational levels.

**Contact email:** kokkinouxa@gmail.com